\documentclass[10pt,twocolumn,letterpaper]{article}

\usepackage[pagenumbers]{cvpr} 

\usepackage[dvipsnames]{xcolor}

\usepackage{amsmath,amsfonts,bm}

\def\eqref#1{equation~\ref{#1}}

\def\1{\bm{1}}

\def\vmu{{\bm{\mu}}}

\def\va{{\bm{a}}}

\def\ve{{\bm{e}}}

\def\vh{{\bm{h}}}

\def\vp{{\bm{p}}}

\def\vs{{\bm{s}}}

\def\vv{{\bm{v}}}
\def\vw{{\bm{w}}}
\def\vx{{\bm{x}}}

\def\vz{{\bm{z}}}

\def\mD{{\bm{D}}}

\def\mI{{\bm{I}}}

\def\mW{{\bm{W}}}

\DeclareMathAlphabet{\mathsfit}{\encodingdefault}{\sfdefault}{m}{sl}
\SetMathAlphabet{\mathsfit}{bold}{\encodingdefault}{\sfdefault}{bx}{n}

\usepackage[accsupp]{axessibility} 

\definecolor{cvprblue}{rgb}{0.21,0.49,0.74}
\usepackage[pagebackref,breaklinks,colorlinks,citecolor=cvprblue]{hyperref}

\def\paperID{10913} 
\def\confName{CVPR}
\def\confYear{2024}

\title{LASIL: Learner-Aware Supervised Imitation Learning For Long-term Microscopic Traffic Simulation}

\author{Ke Guo$^{1,2,3}$, Zhenwei Miao$^{1}$\thanks{Zhenwei Miao is the corresponding author.}, Wei Jing$^{1}$, Weiwei Liu$^{1}$, 
Weizi Li$^{4}$, Dayang Hao$^{1}$, Jia Pan$^{2,3}$ 
\\
$^1$Alibaba Group \quad $^2$The University of Hong Kong \\ \quad $^3$Centre for Transformative Garment Production, Hong Kong \quad $^4$University of Tennessee, Knoxville \\
{\tt\small u3006612@connect.hku.hk, zhenwei.mzw@alibaba-inc.com}\\
{ \tt\small 21wjing@gmail.com, 11932061@zju.edu.cn, weizili@utk.edu,  jpan@cs.hku.hk}
}

\begin{document}
\maketitle

\begin{abstract}
Microscopic traffic simulation plays a crucial role in transportation engineering by providing insights into individual vehicle behavior and overall traffic flow. However, creating a realistic simulator that accurately replicates human driving behaviors in various traffic conditions presents significant challenges. Traditional simulators relying on heuristic models often fail to deliver accurate simulations due to the complexity of real-world traffic environments. Due to the covariate shift issue, existing imitation learning-based simulators often fail to generate stable long-term simulations. In this paper, we propose a novel approach called learner-aware supervised imitation learning to address the covariate shift problem in multi-agent imitation learning. By leveraging a variational autoencoder simultaneously modeling the expert and learner state distribution, our approach augments expert states such that the augmented state is aware of learner state distribution. Our method, applied to urban traffic simulation, demonstrates significant improvements over existing state-of-the-art baselines in both short-term microscopic and long-term macroscopic realism when evaluated on the real-world dataset pNEUMA.
\end{abstract}

\section{Introduction}

Microscopic traffic simulation is a cornerstone in transportation engineering. It enables engineers to predict and analyze individual vehicle behavior, providing crucial insights into how alterations in road structures or traffic management strategies might influence overall traffic flow; it allows for the testing of diverse scenarios without disrupting real-world traffic; and it enhances safety by pinpointing potential hazards and devising strategies to mitigate risks.
By leveraging simulations, engineers can optimize traffic flow, reduce congestion, and enhance overall efficiency, proving particularly advantageous in designing or enhancing road networks. 
Using the simulation data, policymakers and urban planners can make  informed decisions aligned with community needs regarding transportation infrastructure.

However, generating realistic and accurate simulations that can simultaneously replicate the microscopic response of human drivers in various traffic conditions and long-term macroscopic traffic statistics is challenging.
In recent years, significant efforts have been invested in developing realistic traffic simulators with the goal to accurately model human driving behaviors. Traditional traffic simulators, such as SUMO~\citep{krajzewicz2012sumo}, AIMSUN~\citep{barcelo2005AIMSUN}, and MITSIM~\citep{yang1996microscopic}, typically rely on heuristic car-following models like the Intelligent Driver Model (IDM)~\citep{treiber2000idm}. 
However, despite carefully calibrating parameters, the rule-based models often fail to deliver accurate simulations due to the complexity of real-world traffic environments~\citep{Feng2021IntelligentDI,liu2023traco}. Factors such as the road structure, neighboring vehicles, and even driver psychology can influence the decisions of human drivers, making it challenging to achieve accurate simulations~\cite{Wilkie2015Virtual,Li2017CityFlowRecon,chao2020survey}.

In pursuit of realistic traffic simulations, researchers have turned to neural networks to represent the driving model through imitation learning (IL) from human demonstrations. Most traffic simulation approaches~\cite{morton2016analysis,Bergamini2021SimNetLR,Suo2021TrafficSim} leverage behavior cloning (BC)~\cite{Michie1990BC} to learn a driving policy by minimizing the disparity between the model output and the human demonstrations in training data. However, BC is hindered by \textit{covariate shift}~\cite{ross2011reduction}, where the state induced by the learner's policy progressively diverges from the expert's distribution. Existing BC-based simulators have succeeded in short-term (less than 20 seconds) simulation applications like autonomous driving tests but often fail to generate stable long-term simulations. 

To address covariate shift, existing methods such as DAgger~\citep{ross2011reduction}, DART~\citep{laskey2017dart}, and ADAPS~\cite{Li2019ADAPS} incorporate supervisor (humans or principled simulations) corrections at the learner's or perturbed expert's states. 
However, human supervision can be problematic due to intensive labor and judgment errors, and principled simulations may not account for heterogeneous driving behaviors.  
Recent traffic simulators~\cite{song2018multi,Zheng2021ObjectiveawareTS} propose using inverse reinforcement learning (IRL). These methods, such as generative adversarial imitation learning (GAIL)~\cite{ho2016generative} and adversarial inverse reinforcement learning (AIRL)~\cite{Fu2018LearningRR}, learn a reward function using a discriminator neural network within Generative Adversarial Networks (GANs)~\cite{Goodfellow2014GenerativeAN}. 
The policy network is trained to maximize the learned reward through online reinforcement learning (RL), enabling agents to handle out-of-distribution states. However, directly applying GAIL to traffic simulation can be problematic~\cite{Bhattacharyya2019SimulatingEP}. The dynamic nature of the environment in the multi-agent system can lead to noises during policy learning, resulting in highly biased estimated gradients. 
Furthermore, training the discriminator in GAIL is challenging due to the instability and sensitivity of hyperparameters during min-max optimization. 

To address the issue of \textit{covariate shift} in multi-agent imitation learning without depending on costly expert supervision or unstable discriminators and multi-agent RL, we propose \textbf{Learner-Aware Supervised Imitation Learning} (LASIL) for tackling the covariate shift problem during policy learning. We mitigate the distribution shift between expert and learner state distribution by augmenting the expert state distribution. However, there is no expert supervision at any augmented state, so we ensure that an augmented state is close to an expert state such that the future trajectory of the expert state can serve as the target trajectory for the augmented state to constrain the learner within the expert state distribution. Hence, our goal is to augment the expert state to cover the learner's state distribution while remaining close to the original expert state distribution. To achieve this, we use a variational autoencoder (VAE)~\cite{2014Auto} to simultaneously model the distributions of both the expert and learner states. 
By minimizing the VAE's latent space regularization loss of modeling both distributions, we project the expert and learner states into a unified latent space. And by minimizing the VAE's reconstruction loss, the resulting reconstruction leveraging such learnt latent space will resemble both expert and learner state distribution. As a result, when inputting expert states into the trained VAE, we obtain a \textbf{learner-aware augmented expert state}. 

In practice, we divide an agent's state into two parts: past trajectory and context, i.e., other features in the state including vehicle type, waypoints, and destination. 
We observe that the distribution of the context is consistent regardless of a learnt policy due to the static features of traffic conditions embedded in the context, leading to less covariate shift. Therefore, we propose a \textbf{context-conditioned VAE} to model the context-conditioned trajectory distribution of both expert and learner states. The decoder of the context-conditioned VAE will receive both the latent variable and the context, and yield only the trajectory information.  

In summary, our contributions are as follows:
\begin{itemize}
    \item We propose learner-aware supervised imitation learning (LASIL) to achieve stable learning and alleviate covariate shift in multi-agent imitation learning.
    \item We propose a learner-aware data augmentation method based on a context-conditioned VAE that generates learner-aware augmented expert states. 
    \item Our approach is tailored for urban traffic simulation. To the best of our knowledge, it is the first imitation learning-based traffic simulator that can reproduce long-term (more than 10 minutes) stable microscopic simulation, achieving 40x simulation length improvements over previous state-of-the-art~\cite{Zheng2021ObjectiveawareTS,Bergamini2021SimNetLR,Suo2021TrafficSim}.
    \item We evaluate our method on the real-world dataset pNEUMA~\cite{barmpounakis2020pNEUMA} with over half a million trajectories. Our approach outperforms state-of-the-art baselines in both short-term microscopic and long-term macroscopic simulations. 
    The code is available at \url{https://github.com/Kguo-cs/LSAIL}.
\end{itemize}

\section{Related Work}

\subsection{Imitation learning}

Existing imitation learning (IL) methods can be broadly categorized into behavior cloning (BC) and inverse reinforcement learning (IRL) approaches. BC~\citep{Michie1990BC} learns a policy in a supervised fashion by minimizing the discrepancy between the learner's actions and those of an expert. However, BC suffers from the issue of covariate shift, where the state distribution induced by the learner's policy gradually deviates from that of the expert. To address this, methods like DAgger~\citep{ross2011reduction} and DART~\citep{laskey2017dart} request supervisor corrections at the learner's or perturbed expert's states. Our method follows a similar supervised learning approach to DAgger and DART, but does not require access to an expert policy.

Due to the challenges in obtaining expert supervision, recent IRL-based methods utilize feedback from a neural network-based discriminator to handle covariate shift. Typically, these methods involve an iterative process alternating between reward estimation and reinforcement learning. Earlier IRL-based methods~\cite{ng2000algorithms,ziebart2008maximum,Shen2022IRL,Guo2022EndtoEndTD} require frequent dynamic programming processes, while recent adversarial IL approaches integrate reward function learning with policy learning using a GAN formulation. However, both GAN and RL training processes are known to be unstable, sensitive to hyperparameters, and have poor sample efficiency~\cite{Dadashi2020PrimalWI}. Moreover, the discriminator can easily exploit insignificant differences between expert and policy samples, leading to undesirable performance~\cite{Zhang2022AutoEncodingAI}. In contrast, our method avoids the min-max optimization problem and the sample-inefficient RL process, requiring minimal fine-tuning. Instead, we utilize the real future trajectory as the target state to learn corrective behavior.

\subsection{Imitation Learning-based Traffic simulation}

Recent traffic simulators focus on enhancing realism by leveraging imitation learning (IL) from human driving demonstrations, which extend traditional BC and IRL methods to tackle the challenging multi-agent IL problem.

BC-based traffic simulators like TrafficSim~\cite{Suo2021TrafficSim} and SimNet~\cite{Bergamini2021SimNetLR} typically begin by training a prediction model and subsequently adjust the predicted trajectories to prevent collisions and adhere to traffic regulations during simulation. However, BC-based methods face challenges in achieving long-term simulation due to the covariate shift problem. To mitigate this issue, we augment expert state based on the policy distribution, enabling stable long-term simulation. Additionally, we enhance performance by modifying the predicted trajectory during simulation through road projection and ensuring smoothness from the current state. Notably, we skip the computationally intensive collision removal operation, prioritizing the capture of macroscopic long-term influence over micro-level details.

IRL-based simulators learn the underlying reward function of human driving behavior and derive the driving policy by maximizing the learned reward. While adversarial IRL methods theoretically address the covariate shift of BC in a single-agent context through online interaction, its performance deteriorates in the multi-agent IL domain due to the dynamic environment, complicating the training process. To tackle this challenge, approaches like PS-GAIL~\cite{song2018multi} and PS-ARIL~\cite{Zheng2021ObjectiveawareTS} adopt two-stage learning and gradually introduce vehicles to the environment. Nonetheless, these methods still exhibit significant undesirable traffic phenomena, such as off-road driving, collisions, and abrupt braking. Building upon PS-GAIL, the reward-augmented imitation learning (RAIL) method~\cite{Bhattacharyya2019SimulatingEP,Koeberle2022ExploringTT} penalizes undesirable phenomena by introducing a hand-crafted reward. However, maximizing the new reward does not guarantee the recovery of human-like trajectories. Despite numerous enhancements to the original GAIL framework, these IRL-based methods often struggle to produce stable long-term traffic flow, as evidenced in our experiments. In contrast, our method is a supervised learning approach, leading to faster, simpler, and more stable learning of driving policies.

\section{Background}
\label{sec:background}
\subsection{Markov Decision Process}

We model the human driving process using a Markov decision process (MDP) denoted as $\mathcal{M}=\{\mathcal{S}, \mathcal{A}, \mathcal{T},r\}$, incorporating a time horizon of $T$, where $\mathcal{S}$ and $\mathcal{A}$ represent the continuous state and action spaces respectively. A stochastic function $\mathcal{T}: \mathcal{S} \times \mathcal{A} \times \mathcal{S} \rightarrow [0,1]$ describes the system dynamics, and $r:\mathcal{S} \times \mathcal{A} \rightarrow \mathbb{R}$ is a reward function. The policy $\pi(\va|\vs)$ determines the probability of selecting an action $a$ at a state $s$, and a trajectory $\tau=(\vs_0,\va_0,\ldots,\vs_T,\va_T)$ represents a sequence of state-action pairs. The marginal state distribution of a policy $\pi$ is computed as $\rho^\pi(\vs)=\frac{1}{D}\sum_{\vs\in \mathcal{D}_\pi}\delta_{\vs}$, where $\mathcal{D}_\pi$ denotes the set of states of size $D$ in trajectories induced by the policy $\pi$, and $\delta_{\vs}$ signifies a Dirac distribution centered on $\vs$. Similarly, the marginal state-action distribution $\rho^\pi(\vs,\va)$ is computed.

\subsection{Imitation Learning}

In the application of IL for learning a driving policy, we assume that all agents adopt the same policy. The objective is to learn a policy $\pi$ that minimizes the $f$-divergence between the marginal state-action distribution of the expert's demonstrations $\rho^{\mathrm{exp}}(\vs,\va)$ and the learner policy's distribution $\rho^{\pi}(\vs,\va)$. For example, BC optimizes the policy to minimize the Kullback-Leibler (KL) divergence at the expert state distribution $\mathbb{E}_{\rho^\mathrm{exp}(\vs)}[\mathrm{KL}(\pi^{\mathrm{exp}}(\va|\vs)\|\pi(\va|\vs))]$, while DAgger minimizes $\mathbb{E}_{\rho^\mathrm{\pi}(\vs)}[\mathrm{KL}(\pi^{\mathrm{exp}}(\va|\vs)\|\pi(\va|\vs))]$. On the other hand, GAIL minimizes the Jensen-Shannon divergence $D_\mathrm{JS}(\rho^{\mathrm{exp}}(\vs,\va)\|\rho^\pi(\vs,\va))$, while AIRL minimizes the KL divergence $\mathrm{KL}(\rho^{\pi}(\vs,\va)\|\rho^{\mathrm{exp}}(\vs,\va))$.

However, GAIL and AIRL often exhibit unsatisfactory practical performance due to their optimization processes' instability and sample inefficiency, which involve GANs and RL. In contrast, BC only requires simple and stable supervised learning but suffers from the covariate shift issue, as it only minimizes the policy difference at the expert state distribution without guaranteeing performance at the learner state distribution. DAgger addresses the covariate shift problem but requires access to the expert policy. To mitigate the covariate shift problem without depending on the expert policy, we propose maximizing the transition probability $\mathbb{E}_{\vs \sim \rho^\mathrm{\exp}(\vs), \vs+\boldsymbol{\epsilon} \sim \rho^\mathrm{\pi}(\vs)}[\mathcal{T}(\vs+\boldsymbol{\epsilon}, \pi(\vs+\boldsymbol{\epsilon}),\vs')]$, where $\boldsymbol{\epsilon}$ is a small augmentation term. Note that our policy learns to predict the distribution of the next state $\vs'$ instead of the action. Our approach assumes that when the expert's original future trajectory can serve as supervision to guide the agent back towards the expert distribution. 

\subsection{Variational Autoencoder}

The VAE defines a generative model given by $p_\theta(\vx,\vz)=p(\vz)p_\theta(\vx|\vz)$, where $\vz$ is the latent variable with prior distribution $p(\vz)$, and $p_\theta(\vx|\vz)$ represents the conditional distribution modeling the likelihood of data $\vx$ given $\vz$. The learning objective is to maximize the training samples' marginal log-likelihood $\log p_\theta(\vx)$. However, due to the intractability of marginalization, VAE maximizes the variational lower bound using $q_\phi(\vz|\vx)$ as the approximate posterior:
\begin{equation}
\begin{split}
&\log p_\theta(\vx) \\
& \geq \mathbb{E}_{\vz \sim q_\phi(\vz|\vx)}\left[\log p_\theta(\vx|\vz)\right]-\mathrm{KL}\left(q_\phi(\vz|\vx) \| p(\vz)\right) \\
& := -\mathcal{L}_{\mathrm{rec}}(\vx)-\mathcal{L}_{\mathrm{KL}}(\vx),
\end{split}
\end{equation}
where $\mathcal{L}_{\mathrm{rec}}(\vx)$ represents the reconstruction loss, which penalizes the network for creating outputs different from the input. $\mathcal{L}_{\mathrm{KL}}(\vx)$ represents the KL divergence loss to make a continuous and smooth latent space, allowing easy random sampling and interpolation. Intuitively, this KL loss encourages the encoder to distribute all encodings evenly around the center of the latent space.

\section{Method}

\begin{figure*}[t]
	\centering
	\includegraphics[width=\linewidth]{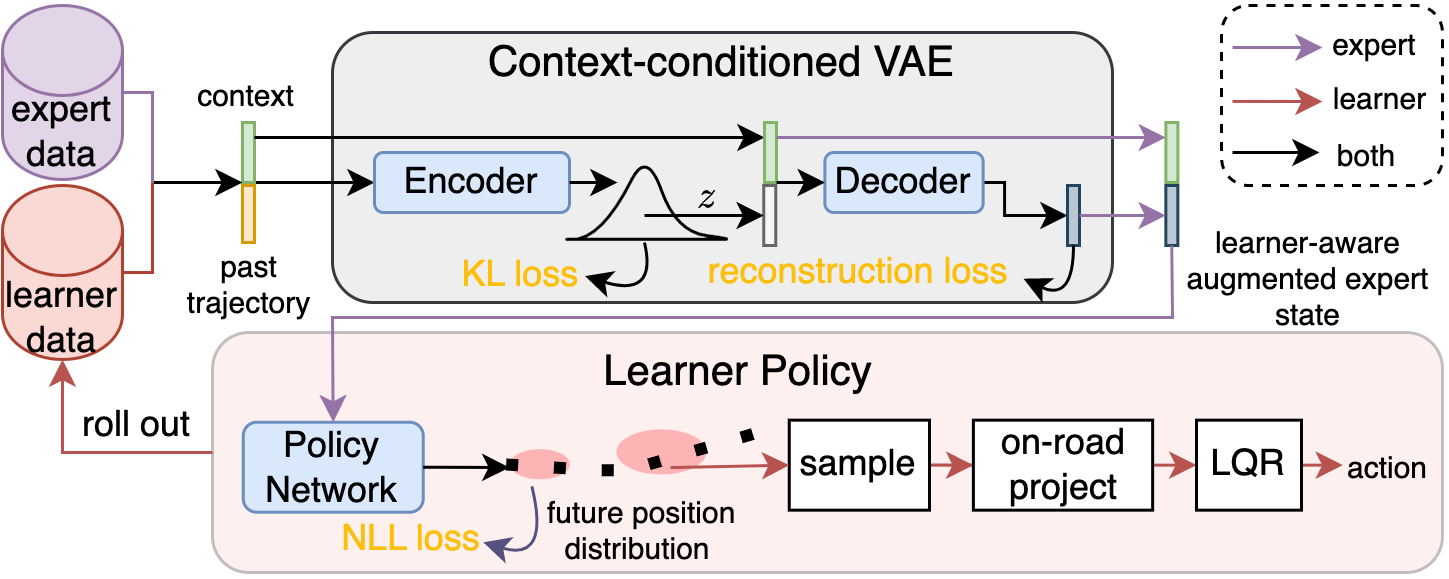}
	\vspace{-0.8cm}
	\caption{Overview of our approach. The processes with purple and red arrows handle only expert or learner data, respectively, while the processes represented by black arrows apply to both data. Each state is a multi-agent state represented by a graph. Each model, including the VAE encoder, decoder, and policy network, is implemented using an EGAT network.}
 	\vspace{-0.6cm}
	\label{fig:overview}
\end{figure*}

In~\cref{fig:overview}, we present an overview of our method. Our model comprises three modules: a VAE-based data augmentation module, a policy network, and a post-processing (LQR and on-road projection) module. During training, we augment each expert data by generating a learner-aware augmented expert state through the VAE's reconstruction of its past trajectory. Using this augmented state along with the original future trajectory, we train the learner's policy network through supervised learning. During simulation, we roll out the policy network with several post-processing steps including sampling, projecting the trajectories onto the road, and smoothing the projected trajectories.

\subsection{State Representation}

As human drivers make decisions mainly depending on their surrounding information, we build a graph to model the traffic system with each driving agent as its node. 

\textbf{Node:} The state of each agent can be divided into two components: the past trajectory $\vs_p$ and the context $\vs_c$. In practice, the past trajectory is composed of the agent's positions at several past time steps, including the current time step. The context includes its type, nearby waypoints' positions and corresponding road width, the traffic light status of the road it is traveling on, and destination. The waypoints are composed of the center points of the routing roads with a fixed interval. We transform each agent's state into its individual coordinate system to learn a policy with transformation invariance. To reduce the implicit covariate shift, we set each agent's current position adding a Gaussian perturbation as the origin and the $x$-axis directions pointing towards the agent's destination like~\cite{Guo2023CCIL}.  

\textbf{Edge:} When each agent's state coordinates are transformed from the global coordinate system to their individual coordinate system, information about the relative positions among agents is lost. However, a traffic model needs the relative information of agents to understand how they interact with each other. To preserve these relationships, we introduce directed edges between neighboring agents. The edge feature is the relative position of the destination node's coordinate origin in the source node's coordinate system. 

\subsection{Learner-aware Data Augmentation}

To address the covariate shift issue between the expert and learner state distributions, we propose to minimize the expert's and learner's embedded state distribution difference. In practice, we propose utilizing the same VAE to model the expert and learner state distributions simultaneously. As both the expert and learner state are projected to the same latent space, the distribution of the state reconstructed from the joint latent space can resemble both distributions. 

In contrast to the past trajectory, the context distribution is more challenging to model but exhibits less covariate shift. Therefore, we propose using a context-conditioned VAE to specifically model the context-conditioned trajectory distribution rather than the state distribution. For each expert or learner state represented as $\vs = (\vs_p,\vs_c)$, we employ an encoder $q_\phi(\vz|\vs_p,\vs_c)$ to obtain the latent variable distribution. We can sample a latent variable $\vz$ from this distribution by applying the reparameterization trick. Subsequently, a decoder $p_\theta(\vs_p|\vz,\vs_c)$ is used to reconstruct the distribution of the past trajectory $\vs_p$ given $\vz$ and $\vs_c$. The VAE is trained to maximize the variational lower bound, which incorporates the context-conditioned log-likelihood of both the expert and learner past trajectories:
\begin{equation}
\mathcal{L}_{VAE} = \mathbb{E}_{\vs \sim \rho^{\exp}}\left[\mathcal{L}(\vs_p|\vs_c)\right] + \lambda \mathbb{E}_{\vs \sim \rho^{\pi}}\left[\mathcal{L}(\vs_p|\vs_c)\right].
\end{equation}
Here, $\lambda$ is a hyperparameter that controls the degree to which the augmented context-conditioned past trajectory distribution aligns with the learner's context-conditioned past trajectory distribution. The term $\mathcal{L}(\vs_p|\vs_c)$ is given by:
\begin{equation}
\begin{split}
\mathcal{L}(\vs_p|\vs_c) = &\mathbb{E}_{\vz \sim q_\phi(\vz|\vs_p,\vs_c)}\left[\log p_\theta(\vs_p|\vz,\vs_c)\right]\\
&- \mathrm{KL}\left(q_\phi(\vz|\vs_p,\vs_c) \| p(\vz)\right).
\end{split}
\end{equation}
For simplicity, we assume $p_\theta$ and $q_\phi$ as multivariate normal distributions with diagonal variance matrices. The prior distribution $p(\vz)$ is set as an isotropic unit Gaussian $\mathcal{N}(\mathbf{0},\mathbf{I})$.

\subsection{Edge-enhanced Graph Attention Network}

We build all models in our approach including the VAE encoder, decoder, and learner's policy network based on an edge-enhanced graph attention (EGAT) network~\cite{diehl2019graph,Mo2022MultiAgentTP}, which model interactions by aggregating neighboring node and edge information using an attention mechanism. The node features in the first layer are obtained by embedding the node input with a fully connected layer and the node features in the other layers are calculated by:
\begin{equation}
    \vh_i^l=\sigma \left(\sum_{j \in \mathcal{N}_i} \alpha_{i j}^l \mW^l \left[\vh^{(l-1)}_i \|  \ve_{i j}\|\vh^{(l-1)}_j \right]\right),
\end{equation}
where $\vh_i^l$ is the feature of node $i$ in the $l$ th layer, $\ve_{ij}$ is the coordinate origin's position of node $j$ relative to node $i$, $\|\|$ means the concatenation, $\mW^l$ is the learnable weight matrix, $\mathcal{N}_i$ is the set of the first-order neighbors of node $i$ (including the node itself), and $\sigma$ is a non-linear activation function. The attention coefficient $\alpha_{ij}^l$ indicates the importance of node $j$ to node $i$, considering both node and edge features, and is computed as:
\begin{equation}
\alpha_{i j}^l=\operatorname{softmax}_j \left(\sigma\left((\vw^l)^T\left[\vh^{(l-1)}_i \|  \ve_{i j}\|\vh^{(l-1)}_j \right]\right)\right),
\end{equation}
where $\vw^l$ is a learnable weight vector, and normalization is performed on the weights across all neighbors of node $i$ using a $\operatorname{softmax}$ function. After passing through multiple EGAT layers, the node features at the last layer is fed into a fully connected layer to obtain the outputs of the network.

\subsection{Policy Loss}

To train the learner's policy network, we first sample a \textbf{learner-aware augmented expert state} from the \textbf{context-conditioned VAE} for each expert state. Then, the policy network predicts its future position distribution over $T$ time steps, denoted as $p(\hat{\vp}_1^m,\hat{\vp}_2^m,...,\hat{\vp}_T^m)$, which is assumed to be a product of multi-variable Gaussian distributions:
\begin{equation}
p(\hat{\vp}_1^m,\hat{\vp}_2^m,...,\hat{\vp}_T^m)=\prod_{t=1}^{T} \mathcal{N}(\hat{\vmu}^m_t, \hat{\boldsymbol{\Sigma}}^m_t),
\end{equation}
where $\hat{\vmu}^m_t$ and $\hat{\boldsymbol{\Sigma}}^m_t$ represent the mean and covariance matrix of the predicted position $\hat{\vp}_t^m$ at future time step $t$. We assume that there is no correlation between the position distributions at different future time steps. To learn the policy network, we minimize the negative log-likelihood (NLL) loss of all agents' ground-truth future trajectories:
\begin{equation}
    \mathcal{L}_{NLL}=-\sum_{m=1}^{M}\sum_{t=1}^{T} \log(\mathcal{N}(\vp_t^{m}-\hat{\vmu}^m_t, \hat{\boldsymbol{\Sigma}}^m_t)) ,
\end{equation}
where $\vp_t^{m}$ denotes the ground truth position of agent $m$ at future time step $t$, and $M$ is the total number of agents.

\subsection{Simulation Process}

During training, we simultaneously simulate (roll out) the learned policy network iteratively to get learner state samples. Instead of directly updating each agent to its predicted position, we apply several post-processing steps to the prediction for better realism. Firstly, we sample from the distribution, and then project each sampled position onto the nearest on-road point. Then, we smooth the projected trajectory with a linear-quadratic regulator (LQR)~\cite{aastrom2021feedback} by minimizing the total commutative quadratic cost of a linear dynamic system described by:
\begin{equation}
\left[\begin{array}{c}
    \tilde{\vp}^m_{t+1} \\
    \tilde{\vv}^m_{t+1} \\
    \end{array}\right] 
    =\left[\begin{array}{ccc}
            \mI & \mD \\
            0 & \mI\\
            \end{array}\right] 
            \left[\begin{array}{c}
                \tilde{\vp}^m_t \\
                \tilde{\vv}^m_{t} \\
                \end{array}\right] 
            + \left[\begin{array}{c}
                \mD^2 \\
                \mD
                \end{array}\right] \tilde{\va}^m_t,
\end{equation}
where $\mD$ is a diagonal matrix with the interval of each time step as diagonal entries, and $\tilde{\vp}_t^m$, $\tilde{\vv}^m_t, \tilde{\va}^m_t$ represent the LQR-planned position, velocity, and acceleration. The system is subject to a quadratic cost function given by:
\begin{equation}
    \mathcal{J}=\sum_{m=1}^{M}\sum_{t=1}^{T} \|\tilde{\vp}_t^m-\bar{\vp}^m_t\|^2+\eta_{\va}\|\tilde{\va}_t^m\|^2,
\end{equation}
where the projected predicted position $\bar{\vp}_{m}^{t}$ is considered as the target pose, and the hyper-parameter $\eta_{\va}$ is used to penalize high acceleration. Finally, each agent is updated to the first position of the planned trajectory.

\subsection{Training Process}

At each training step, the policy network is trained to maximize the probability of the expert's future trajectory using its context state and history trajectory augmented by the VAE as input. Simultaneously, the VAE is iteratively trained to reconstruct the expert and learner data. The expert data are the same as the policy network's training data before augmentation. The learner data is sampled from a replay buffer. Every $N$ training steps, we empty the replay buffer and roll out the current policy with post-processing for $S$ time steps to generate the learner data and store it in the replay buffer. During the roll-out, we start at a random time step in the training dataset and apply our model in a closed-loop manner. At each time step during the roll-out, we sample a future trajectory for each vehicle from the Gaussian distribution predicted by our policy network, project the predicted trajectory onto the road, smooth the on-road trajectory with LQR, and finally update the vehicle to the first position of the smoothed trajectory.

\section{Experiment}

\subsection{Dataset}

We utilize a real-world urban dataset called pNEUMA~\cite{barmpounakis2020pNEUMA}, which was collected by 10 drones in Athens over a span of 4 days. The dataset encompasses over half a million vehicle trajectories within a large area encompassing over 100 km of lanes and approximately 100 intersections. The recordings were conducted at 5 intervals each day, with each period lasting about 15 minutes and a data collection time interval of 0.04 seconds. To enhance computational efficiency, we adopt a time step of 0.4 seconds. The dataset is divided into a training set, comprising recordings from the first 3 days, and a validation/test set, consisting of recordings from the final day. Notably, we choose not to utilize other popular traffic datasets like NGSIM~\cite{Colyar2017ngsim} or HighD~\cite{krajewski2018highd} or autonomous driving datasets like Lyft~\cite{houston2020one} or nuPlan~\cite{Caesar2021nuplan}, as they only encompass samll-scale scenarios. This limitation makes it inadequate for evaluate macroscopic realism.

\subsection{Metrics}

We evaluate the realism of our simulator by measuring the similarity between the simulation result and real data. During evaluation, each vehicle enters the simulator at its first recorded time and position, and is controlled by our simulator to complete its recorded route. When an agent reaches its final recorded position, it is removed from the simulator.

\textbf{Short-term microscopic}: Following~\cite{song2018multi,Bhattacharyya2019SimulatingEP}, we conduct a short-term microscopic evaluation by simulating for 20 seconds from a random time step in the test dataset. We first measure the similarity between the simulated and real data using \textbf{position and velocity RMSE} metrics given by:
\begin{equation}
\operatorname{RMSE}=\frac{1}{T_s}\sum_{t=1}^{T_s} \sqrt{\frac{1}{M}\sum_{m=1}^M \|\vs_t^m-\hat{\vs}_t^m\|^2},   
\end{equation}
where $\vs_t^m$ and $\hat{\vs}_t^m$ were the real and simulated value of the position or velocity of the agent $m$ at time step $t$, respectively. $T_s$ was the total simulated time steps, and $M$ was the total simulated agent number. Besides, we measure the \textbf{minimum Average Displace Error (minADE)} to not penalize reasonable trajectory unlike the real one:
\begin{equation}
    \operatorname{minADE}_N=\frac{1}{M}\sum_{m=1}^M\min_{\hat{\vs}_n} \frac{1}{T_s}\sum_{t=1}^{T_s} \|\vs_t^m-\hat{\vs}_{t,n}^m\|^2,
\end{equation}
where $N=20$ is roll-out times.  We also calculate the \textbf{off-road rate} (the avarage proportion of vehicles that deviate more than 1.5m from the road over all time steps). The common collision rate metric is not used because we focus on the long-term impact and the dataset does not provide accurate vehicle size and heading information.

\textbf{Long-term macroscopic}: we also evaluate our model's long-term macroscopic accuracy on five periods for 800 seconds from its initial recording time. To measure the long-term performance, we use two standard macroscopic metrics for traffic flow data~\cite{Lighthill1955OnKW,Richards1956ShockWO,sewall2010continuum}, namely \textbf{road density and speed RMSE}, in addition to the \textbf{off-road rate}. The density of a road at a time step is calculated by dividing the number of vehicles on the road by its total lane length, assuming that all lanes have the same width. Meanwhile, the road speed is computed as the mean speed of all vehicles on the road. To quantify the similarity between the simulated and ground truth values, we still use RMSE, where the variable $M$ becomes the total number of roads.

\begin{table*}[htbp]
\caption{Comparison with baselines and ablated models on microscopic metrics for 20 seconds.}
\vspace{-0.7cm}
\begin{center}
\begin{tabular}{l|ccccc}
\textbf{Model} & \textbf{Position RMSE(m)} & \textbf{Velocity RMSE(m/s)} & $\operatorname{\textbf{minADE}}_{\textbf{20}}$\textbf{(m)} & \textbf{Off-road(\%)}  \\ \toprule 
SUMO~\cite{krajzewicz2012sumo} & 41.25 & 7.00 & 24.20 & \textbf{0}  \\
BC~\citep{Michie1990BC} & 40.08\textpm 1.61  &  6.74\textpm0.37 & 22.02\textpm3.98 & 33.43\textpm 3.10 \\
MARL~\cite{li2023scenarionet} & 48.66\textpm1.54 & 8.92\textpm0.67 & 40.07\textpm1.26 & 3.70\textpm0.40\\
MARL+BC~\cite{lu2022imitation} & 29.78\textpm0.42 & 4.54\textpm0.14 & 22.07\textpm0.21 & 11.14\textpm1.17 \\
PS-GAIL~\cite{song2018multi} & 34.78\textpm0.43 & 5.33\textpm0.12 & 27.86\textpm0.33 & 7.65\textpm0.32 \\
RAIL~\cite{Bhattacharyya2019SimulatingEP} & 31.62\textpm0.65  & 5.51\textpm0.09 & 24.51\textpm0.41 &  2.57\textpm0.47 \\
\textbf{LASIL (ours)} & \textbf{19.21\textpm 0.44} & \textbf{3.02\textpm0.07} & \textbf{12.79\textpm 0.32} & 0.28\textpm 0.01\\
\midrule
w/o Augmentation & 21.62\textpm0.54 & 3.34\textpm0.14 & 13.81\textpm0.37 & 1.31\textpm0.18  \\
w/o Context-conditioned & 23.03\textpm0.87 & 3.73\textpm0.12 & 14.94\textpm0.43 & 0.51\textpm0.04 \\
w/o On-road Projection & 22.53\textpm0.42 & 3.26\textpm0.10 & 14.42\textpm0.33  & 1.74\textpm0.60\\
w/o LQR & 22.62\textpm0.43 & 4.05\textpm0.15  & 14.02\textpm0.28 & 0.22\textpm0.02 \\
\bottomrule
\end{tabular}
\end{center}
\label{tab:short}
\vspace{-0.6cm}
\end{table*}

\begin{table*}[htbp]
\caption{Comparison with baselines and ablated models on macroscopic metrics for 800 seconds.}
\vspace{-0.7cm}
\begin{center}
\begin{tabular}{l|ccc}
\textbf{Model} & \textbf{Road Density RMSE(veh/km)} & \textbf{Road Speed RMSE(m/s)} & \textbf{Off-road(\%)}  \\ \toprule 
SUMO~\cite{krajzewicz2012sumo} & 52.70 & 5.52 & \textbf{0}  \\
BC~\citep{Michie1990BC} & 61.51\textpm 1.53  &  5.38\textpm0.21 & 42.15\textpm 5.25 \\
MARL~\cite{li2023scenarionet} &  90.59\textpm2.65 & 5.27\textpm0.48 & 4.63\textpm0.64 \\
MARL+BC~\cite{lu2022imitation} & 43.60\textpm2.80 & 4.01\textpm0.13 & 22.73\textpm5.62 \\
PS-GAIL~\cite{song2018multi} & 54.06\textpm1.23 & 4.03\textpm0.05 & 13.24\textpm3.20 \\
RAIL~\cite{Bhattacharyya2019SimulatingEP} & 54.45\textpm1.89 & 3.89\textpm0.11 & 2.92\textpm0.38 \\
\textbf{LASIL (ours)} & \textbf{45.13\textpm0.25} &  \textbf{3.17\textpm 0.14} & 0.34\textpm 0.03 \\
\midrule
w/o Augmentation & 55.65\textpm0.44 & 3.73\textpm0.23 & 9.62\textpm1.40  \\
w/o Context-conditioned & 55.98\textpm0.51 & 4.48\textpm0.19 & 0.64\textpm0.08 \\
w/o On-road Projection& 52.90\textpm0.54 & 3.59\textpm0.20 & 11.33\textpm2.37  \\
w/o LQR & 61.68\textpm0.96 & 3.54\textpm0.16  & 0.47\textpm0.07 \\
\bottomrule
\end{tabular}
\end{center}
\label{tab:long}
\vspace{-0.9cm}
\end{table*}

\subsection{Performance}

We compare our method against state-of-the-art baselines:

\textbf{SUMO}~\cite{krajzewicz2012sumo}: we use the IDM model~\cite{treiber2000idm} as the car-following model and mobil~\cite{kesting2007general} as the lane-changing model. We tune the IDM's parameters for 6 types of vehicle by minimizing the MSE between the IDM calculated acceleration and real acceleration using an Adam optimizer~\cite{Kingma2014AdamAM}.

\textbf{BC}~\cite{Michie1990BC}: we learn our model directly by the BC method.

\textbf{MARL}~\cite{li2023scenarionet}: we train our model using IPPO~\cite{de2020independent} as the Multi-Agent RL algorithm, where the reward function is composed of three parts: a displacement reward (average distance between agent trajectory with GT trajectory), an off-road penalty and a terminal reward.

\textbf{MARL+BC}~\cite{lu2022imitation}: we add a behavior cloning term in the loss function while learning the policy with MARL.

\textbf{PS-GAIL}~\cite{song2018multi}: we let all vehicles share the same policy parameter and critic parameter and learn the policy using reward functions computed by GAIL.

\textbf{RAIL}~\cite{Bhattacharyya2019SimulatingEP}: we learn the model as \textbf{PS-GAIL} but with additional displacement, off-road and terminal rewards.

We train and evaluate each model five times to obtain the mean and standard deviation (std) of various metrics. Note that we do not apply on-road projection and LQR for baselines. We evaluate both short-term and long-term performance, as shown in~\cref{tab:short} and~\cref{tab:long}, respectively. Our method achieves better results than all baselines in terms of position and velocity RMSE, road density and speed RMSE, with minor off-road rate.

\subsection{Ablation Study}

We conducted a series of ablation experiments to assess the individual contributions of crucial components in our approach, whose results are presented in \cref{tab:short} and \cref{tab:long}.

\textbf{Augmentation}: By removing our VAE module and directly learning the original expert state and action, we analyze the importance of our data augmentation technique. The results demonstrate that augmentation plays a vital role in improving both short-term and long-term simulation performance by mitigating covariate shift.

\textbf{Context-conditioned VAE}: To emphasize the significance of modeling the context-conditioned trajectory distribution rather than the whole state distribution, we replaced our context-conditioned VAE with a naive VAE that directly models the expert and learner state distribution. The observed drop in performance demonstrates the challenges in reconstructing the context distribution.

\textbf{On-road projection}: Our ablation study on the on-road projection module aims to show its impact on reducing the off-road rate. The results show that the on-road projection module leads to a notable decrease in the off-road rate and moderate improvements in other performance metrics.

\textbf{LQR}: Removing LQR module allows us to evaluate its effectiveness. While LQR led to a higher short-term off-road rate due to more constrained movement (driving off-road due to inertia), its removal deteriorates other short-term and long-term metrics, because LQR can smooth the trajectory, thus leads to more realistic simulations.

\subsection{Qualitative Result}
\begin{figure*}[t]
	\centering
	\includegraphics[width=0.94\linewidth]{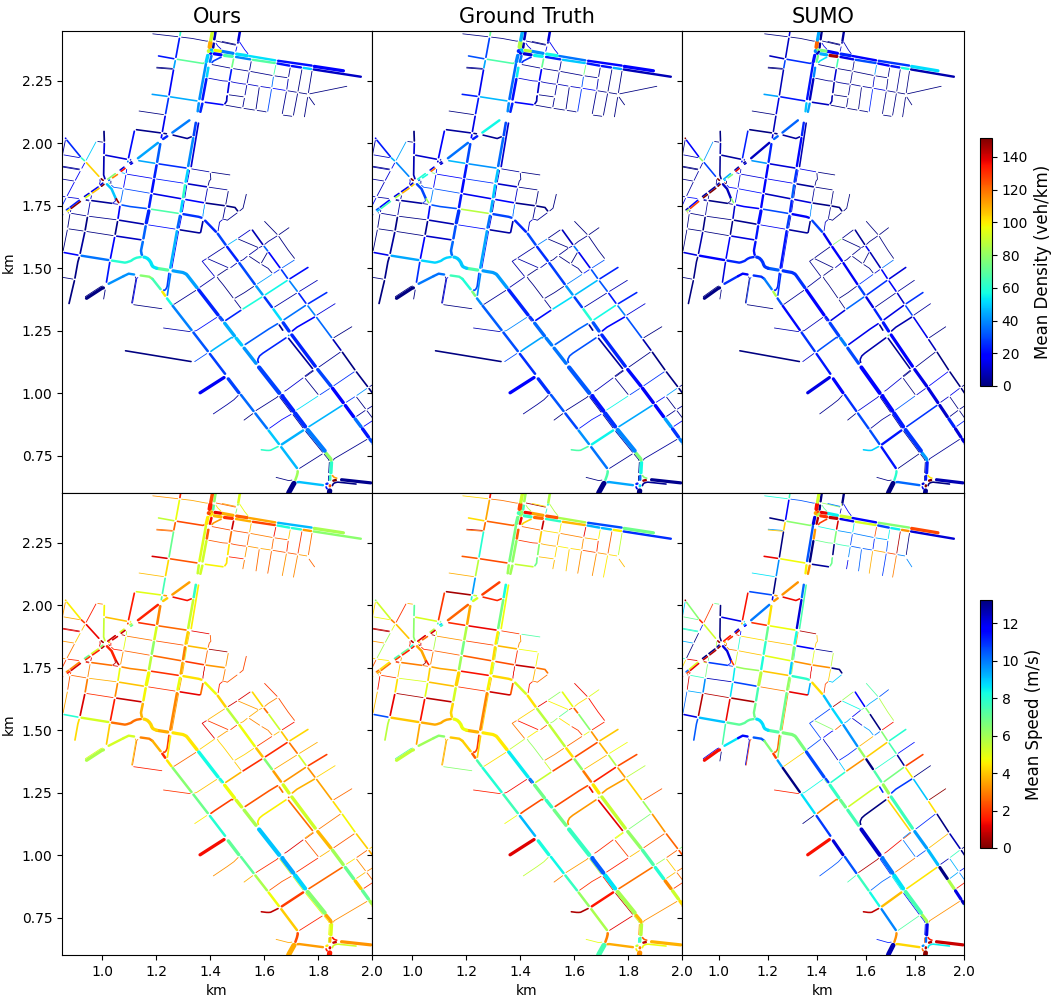}
	\vspace{-0.4cm}
	\caption{Mean density and speed on each road over all time steps in the long-term evaluation. Our method's density and speed hot-maps have a more similar color to the ground-truth one compared with the SUMO's. }
 	\vspace{-0.6cm}

	\label{fig:densityspeed}
\end{figure*}

In~\cref{fig:densityspeed}, we present the mean road density and speed for real-world data, SUMO simulation, and our proposed method over all time steps. The results demonstrate that our proposed method surpasses the capabilities of the SUMO simulator at replicating long-term macroscopic traffic patterns due to our model's enhanced ability to replicate the typical microscopic driving behaviors over a long period.

\section{Conclusion}

In conclusion, we have addressed the challenge of creating a realistic traffic simulator that accurately models human driving behaviors in various traffic conditions. Traditional imitation learning-based simulators often fail to deliver accurate long-term simulations due to the covariate shift problem in multi-agent imitation learning. To tackle this issue, we proposed a learner-aware supervised imitation learning method, which leverages a context-conditioned VAE to generate learner-aware augmented expert states. We leverage a context-conditioned VAE to simultaneously reconstruct the expert and learner state. This approach enables the reproduction of long-term stable microscopic traffic simulations, marking a significant advancement in the field of urban traffic simulation. Our method has demonstrated superior performance over existing state-of-the-art simulators when evaluated on the real-world dataset pNEUMA, achieving better short-term microscopic and long-term macroscopic similarity to real-world data than state-of-the-art baselines. 

\section{Acknowledgements}

This research project is partially supported by the Innovation and Technology Commission of the HKSAR Government under the InnoHK initiative, Hong Kong General Research Fund (11202119 and 11208718), Innovation and Technology Commission (GHP/126/21GD), and Guangdong, Hong Kong and Macao Joint Innovation Project (2023A0505010016).

{
    \small
    \bibliographystyle{ieeenat_fullname}
    \bibliography{main}
}

\clearpage
\setcounter{page}{1}
\maketitlesupplementary

\section{Model details}

The hyper-parameters of our model architecture are listed in~\cref{tab:Hyper-parameters}. All models are trained using the Adam optimizer with a learning rate of 0.0003, and batch size 32 with 8 Ge-force 3090. The VAE are trained simultaneously with the policy network using the online simulated learned data and offline expert data. 

\begin{table}[htbp]
\caption{Hyper-parameters.}
\vspace{-0.3cm}
\begin{center}
\begin{tabular}{l|c}
\textbf{Hyper-parameter} & \textbf{Value}  \\ 
\toprule 
    History time steps &  10 \\
    Future time steps $T$ &  10 \\
    Route point number & 30 \\
    Neighbor number & 6 \\
    Neighbor maximum distance &  20 m\\
    Origin perturbation std & 2 m\\
    EGAT hidden size &  512 \\
    VAE latent dim  & 8 \\
    VAE encoder layer number & 1\\
    VAE decoder layer number & 1\\
    Policy Network layer number & 1\\
    LQR acceleration weight $\eta_\va$ & 1\\
    Learner VAE loss weight $\lambda$ & 1 \\
    Training simulation interval $N$ & 50 \\
    Training simulation length $S$ & 50 \\
\bottomrule
\end{tabular}
\end{center}
\label{tab:Hyper-parameters}
\end{table}

\section{Baseline details}
\textbf{SUMO} uses the mobil model and IDM with various tuned parameters (including desired speed, acceleration, deceleration, minimum gap, time headway) for 6 types (including motorcycle, car, taxi, bus, medium and heavy vehicle) of vehicles. These parameters are shown in~\ref{tab:Sumo-parameters}. 

\textbf{BC}: our model without the VAE, LQR and the on-road projection module. 

All RL baselines are trained using IPPO with default parameters in the Ray library. 
\textbf{MARL}: the reward is the sum of a displacement reward (weight 0.01), an off-road penalty (weight 1), and a terminal reward (weight 0.01). 

\textbf{MARL+BC}: adds a BC term (weight 1) to the loss function of the MARL policy. 

\textbf{PS-GAIL}: learns a policy using reward functions from a discriminator network, which is also trained using Adam with a learning rate of $0.0003$. 

\textbf{RAIL}: PS-GAIL with additional rewards from MARL.

\begin{table*}[htbp]
\caption{Sumo hyper-parameters.}
\vspace{-0.3cm}
\begin{center}
\begin{tabular}{l|cccccc}
\textbf{Hyper-parameter} & \textbf{Motorcycle} & \textbf{Car}  & \textbf{taxi} & \textbf{bus} & \textbf{medium vehicle} & \textbf{heavy vehicle} \\
\toprule 
    desired speed ($m/s$) &  30 & 30 & 30  & 11.70 & 30 & 17.38 \\
    acceleration ($m/s^2$) & 2.5 & 2.5 & 2.5  & 2.5 & 2.5 & 2.5  \\
    deceleration ($m/s^2$) & 10.0 & 10.0 & 10.0  & 10.0 & 10.0 & 10.0 \\
    minimum gap ($s$)& 0.1 & 0.1 & 0.1  & 0.1 & 0.1 & 0.1 \\
    time headway ($s$) &  0.1 &   0.1  &   0.1   &   0.1  &   0.1  &   0.1  \\
\bottomrule
\end{tabular}
\end{center}
\label{tab:Sumo-parameters}
\end{table*}

\section{Dataset Preparation}

The data preparation process of pNEUMA dataset is introduced in this section. 

\subsection{Trajectory Data}

The trajectory data is downloaded from the official website (\url{https://open-traffic.epfl.ch}), specifically the data recorded by ALL Drones during all periods except for the first period (8.30-9.00 at 2018/10/24) due to a large position error caused by wind gusts. 

\subsection{Routing}

To determine the route for each vehicle trajectory, we used the method in \url{https://github.com/wannesm/LeuvenMapMatching}. However, this method generated many circular routes that are rarely observed in real data. To address this issue, we skipped the intermediate routing node points if they were far away from the actual trajectory, which helped reduce the number of unrealistic circular routes.

\subsection{Road Network}

The map information is downloaded from OpenStreetMap, and then we import it into SUMO to generate the road network. We only include highways for vehicles in the road network while excluding other road types, such as sidewalks and railways. However, we find that the map data is not always accurate, so we manually adjust the road shapes to reduce the number of off-road driving cases in the recorded trajectory data. Additionally, we modify the lane connection relations in junctions to alleviate traffic jams during SUMO simulations.

\subsection{Traffic Light}

Because the traffic light information is not provided by the dataset, we design an algorithm to estimate traffic light information from the recorded trajectory data. 

Firstly, we filter all vehicle starting and stopping points near all signaled intersections from the trajectory data.

Secondly, we cluster these points based on their located edge, as we assume that all lanes on one edge are controlled by the same traffic light.

Thirdly, we obtain all time steps when each traffic light turns green by identifying its corresponding clustered points whose time gap to the previous point is larger than seven seconds. Similarly, we can obtain all time steps when it turns red.

Fourthly, based on all the time steps when the traffic light turns green, we need to calculate the traffic light's first turning green time step, green time, and cycle length. We assume that all traffic lights in the same junction have the same cycle length, which can only be 45 or 90 seconds. To estimate the first green time and cycle length, we use a cost function, where a negative cost is given if the filtered turning green time steps match the estimated turning green time step, and a positive cost is given if there is no filtered turning green time step matching the estimated turning green time step. By enumerating all first turning green time steps with an interval of 0.01 seconds, and cycle length of 45 or 90 seconds, we output the result with the minimal costs. Based on the other traffic light turning green time steps in the same junction, we can obtain its turning red time. If there is no other traffic light in the same junction, we need to estimate its turning red time as we do in the estimation of green time. Based on the turning red time and the turning green time, we can obtain the green time of each traffic light.

\section{Runtime}

\begin{figure}[htbp]
 	\caption{Runtime of each time step during the long-term evaluation.}
	\centering
	\includegraphics[width=\linewidth]{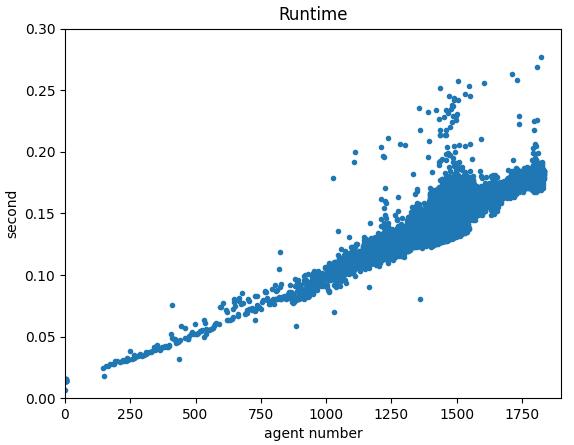}
	\label{fig:Runtime}
\end{figure}

We perform runtime experiments using a single Nvidia GeForce GTX 1080 GPU and an Intel i7-8700@3.2GHz CPU. These experiments take into account all components of our traffic model, including input preparation, trajectory prediction, and action generation. The runtime results for all time step are recorded during the long-term evaluation, as shown in~\cref{fig:Runtime}. We can see that the runtime increases almost linearly with the number of agents. Besides, our method can finish one simulation step of thousands of agents within an acceptable time limit (smaller than 1 second). 

\section{Qualitative results}

\subsection{Prediction and Planned Trajectories}

\begin{figure*}[htbp]
	\centering
	\includegraphics[width=0.95\linewidth]{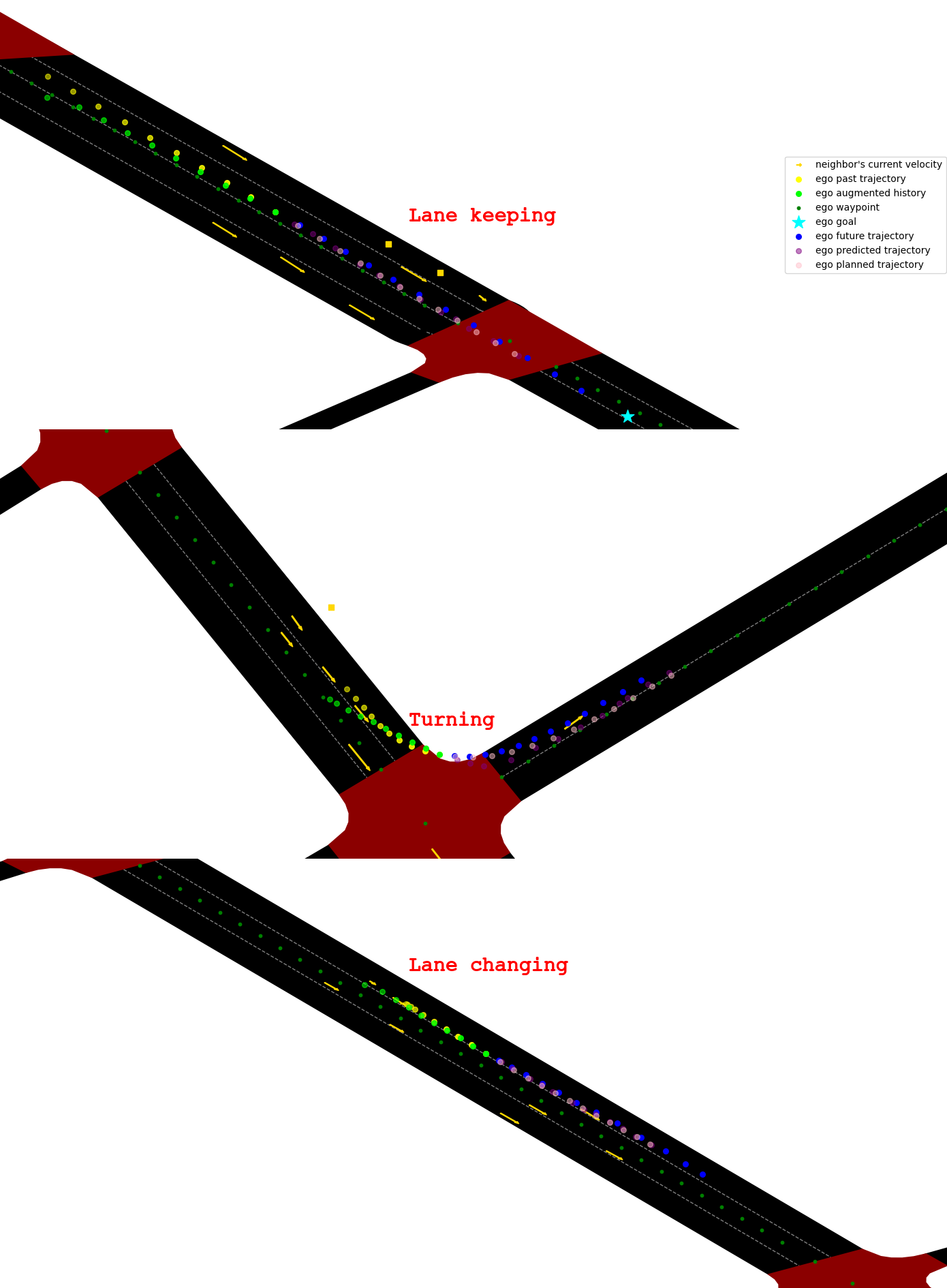}
  	\vspace{-0.3cm}
 	\caption{Trajectories augmented by our context-conditioned  VAE, predicted by our policy network and subsequent planned trajectory by the LQR module in lane keeping, turning and lane changing scenarios.}
	\label{fig:micro}
\end{figure*}

In~\cref{fig:micro}, we present the trajectories produced by our context-conditioned VAE during training. Based on the augmented past trajectory and context information, our policy network predicts a future trajectory, which is subsequently refined by the LQR module, in different scenarios such as lane keeping, turning, and lane changing. The results illustrate that our context-conditioned VAE is capable of generating a wide range of past trajectories that encompass the distribution of possible policies, while remaining reasonable and closely resembling the actual past trajectory. Moreover, our method accurately predicts future trajectories that closely align with the actual path, based on the augmented history and context. Additionally, the incorporation of the LQR module enhances the smoothness of the trajectory. Importantly, our approach also demonstrates the ability to generate diverse behaviors that comply with the surrounding environment.

\subsection{Statistical Distribution}

\begin{figure*}[htbp]
	\centering
	\includegraphics[width=0.88\linewidth]{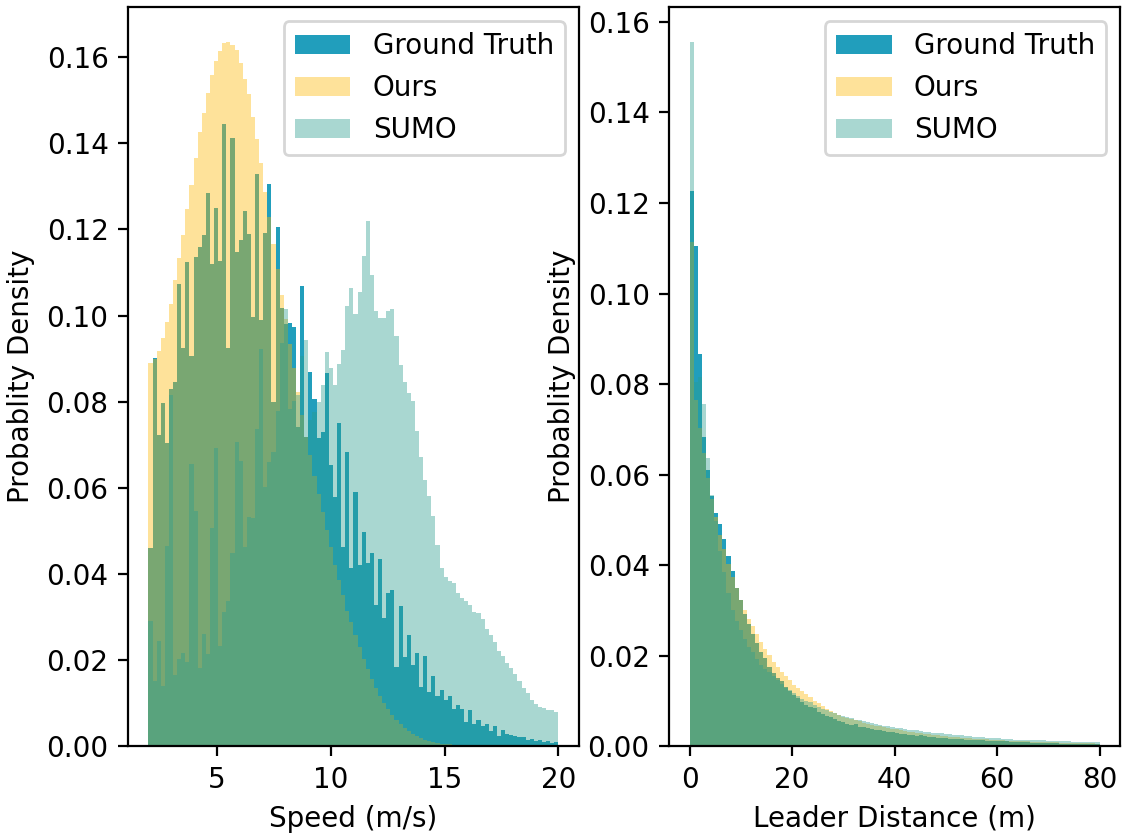}
   	\vspace{-0.3cm}
 	\caption{Distributions of speed and leader vehicle's distance in long-term evaluation.}
	\label{fig:distribution}
  	\vspace{-0.3cm}
\end{figure*}

In~\cref{fig:distribution}, we illustrate the distribution of speed and distance to the leading vehicle during long-term simulations. Our method produces more similar speed distributions to the ground truth than SUMO since the Intelligent Driver Model (IDM) always aims to move at the highest speed. Furthermore, our method generates leader distance distributions that closely match the ground truth.

\section{Road Shape Change Experiment}

\begin{figure*}[t]
	\centering
	\includegraphics[width=0.88\linewidth]{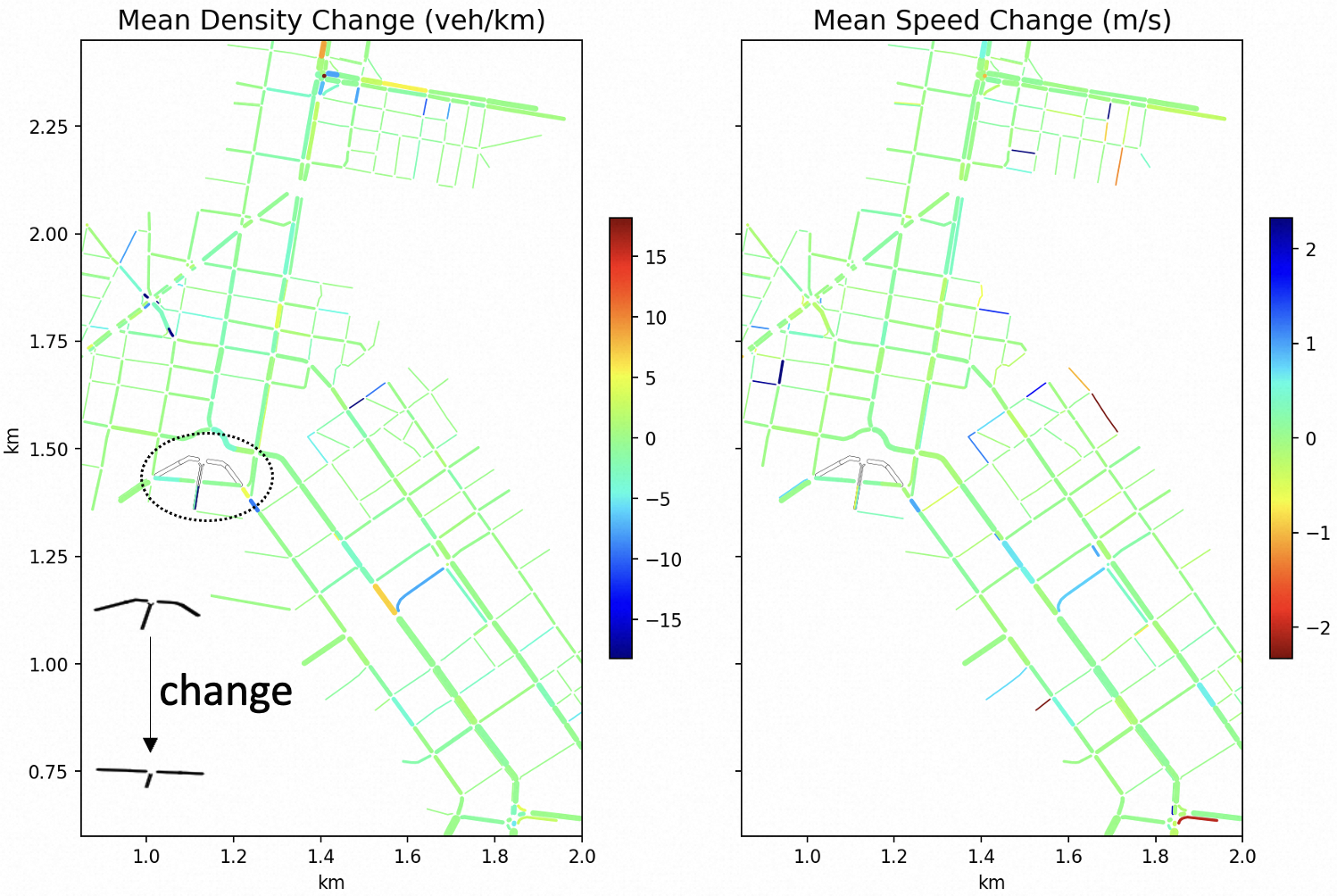}
  	\vspace{-0.3cm}
	\caption{Mean density and speed changes after modifying road network.}
	\label{fig:change}
\end{figure*}

Our microscopic long-term traffic simulators can help transportation engineers and planners to analyze and predict the impact of microscopic adjustments on traffic patterns without disrupting real-world traffic. For example, it can help analyze how changing road shape affects traffic patterns. In~\cref{fig:change}, we present the mean road density and speed changes in our simulator after modifying several roads' shapes. We can see that a local microscopic modification in road network can causes traffic congestion or alleviation in distant areas.

\end{document}